\pdfoutput=1

\documentclass[11pt]{article}

\usepackage[]{acl}
\usepackage{comment}
\usepackage{graphicx}

\usepackage{times}
\usepackage{latexsym}

\usepackage[T1]{fontenc}

\usepackage[utf8]{inputenc}

\usepackage{microtype}

\usepackage{inconsolata}

\usepackage{multirow} 

\newcommand{\forget}[1]{}

\title{MTP: A Dataset for Multi-Modal Turning Points in Casual Conversations}

\usepackage{authblk}

\author{\textbf{Gia-Bao Dinh Ho}$^1$, \textbf{Chang Wei Tan}$^2$, 
        \textbf{Zahra Zamanzadeh Darban}$^2$, \textbf{Mahsa Salehi}$^2$,\\
        \textbf{Gholamreza Haffari}$^2$, \textbf{Wray Buntine}}

\affil[1]{VinUniversity, Ha Noi, Viet Nam}
\affil[2]{Monash University, Melbourne, Australia}

\affil[ ]{\texttt{\{bao.dhg, wray.b\}@vinuni.edu.vn}}
\affil[ ]{\texttt{\{Chang.Tan, Zahra.Zamanzadeh, mahsa.salehi, gholamreza.haffari\}@monash.edu}}

\newtheorem{definition}{Definition}

\begin{document}
\maketitle

\begin{abstract}
Detecting critical moments, such as emotional outbursts or changes in decisions during conversations, is crucial for understanding shifts in human behavior and their consequences.
Our work introduces a novel problem setting focusing on these moments as \textit{turning points (TPs)}, accompanied by a meticulously curated, high-consensus, human-annotated multi-modal dataset. 
We provide precise timestamps, descriptions, and visual-textual evidence highlighting changes in emotions, behaviors, perspectives, and decisions at these turning points. 
We also propose a framework, TPMaven, utilizing state-of-the-art vision-language models to construct a narrative from the videos and large language models
to classify and detect turning points in our multi-modal dataset. Evaluation results show that TPMaven achieves an F1-score of 0.88 in classification and 0.61 in detection, with additional explanations aligning with human expectations.
\end{abstract}

\section{Introduction}
Identifying key moments in videos, like highlight detection or moment retrieval, is crucial. This involves pinpointing moments through scene changes or specific descriptions using matching and strategic comparison processes. Turning point (TP) classification and detection enhance this by incorporating reasoning to identify significant conversational shifts. The challenge lies in the complex reasoning needed, evident in our data annotation where even human annotators require group discussions. Detecting these turning points is vital for post-analysis of conversations, recognizing moments that impact speakers' reactions. Understanding these moments enhances future interactions, particularly valuable in new or unfamiliar settings like therapy or negotiation, and offers strategies for successful outcomes.
Given limitations in existing multi-modal datasets and the novelty of our research, we aim to pioneer the creation of a novel high-quality dataset with turning points. Collecting four seasons of The Big Bang Theory TV series, with its eccentric characters likely causing turning points, we focus on 40 episodes from seasons 1 to 4, specifically on conversations. 

This study makes several contributions: (1) Introducing Multi-modal Turning Point Classification (MTPC), Multi-modal Turning Point Detection (MTPD), and Multi-modal Turning Point Reasoning (MTPR) tasks in human casual conversation.
(2) Curated a human-annotated \textbf{M}ultimodal \textbf{T}urning \textbf{P}oints (MTP) dataset for casual conversation, enriched with textual and visual cues depicting subjective personal states.
(3) Proposing a novel framework for MTPC and MTPD, utilizing vision language models (VLMs) for narrative construction and large language models (LLMs) for effective reasoning in turning point detection.
(4) The code and data are publicly available.\footnote{\url{https://giaabaoo.github.io/TPD_website/}}


\begin{figure*}
  \centering
  \includegraphics[width=0.8\linewidth]{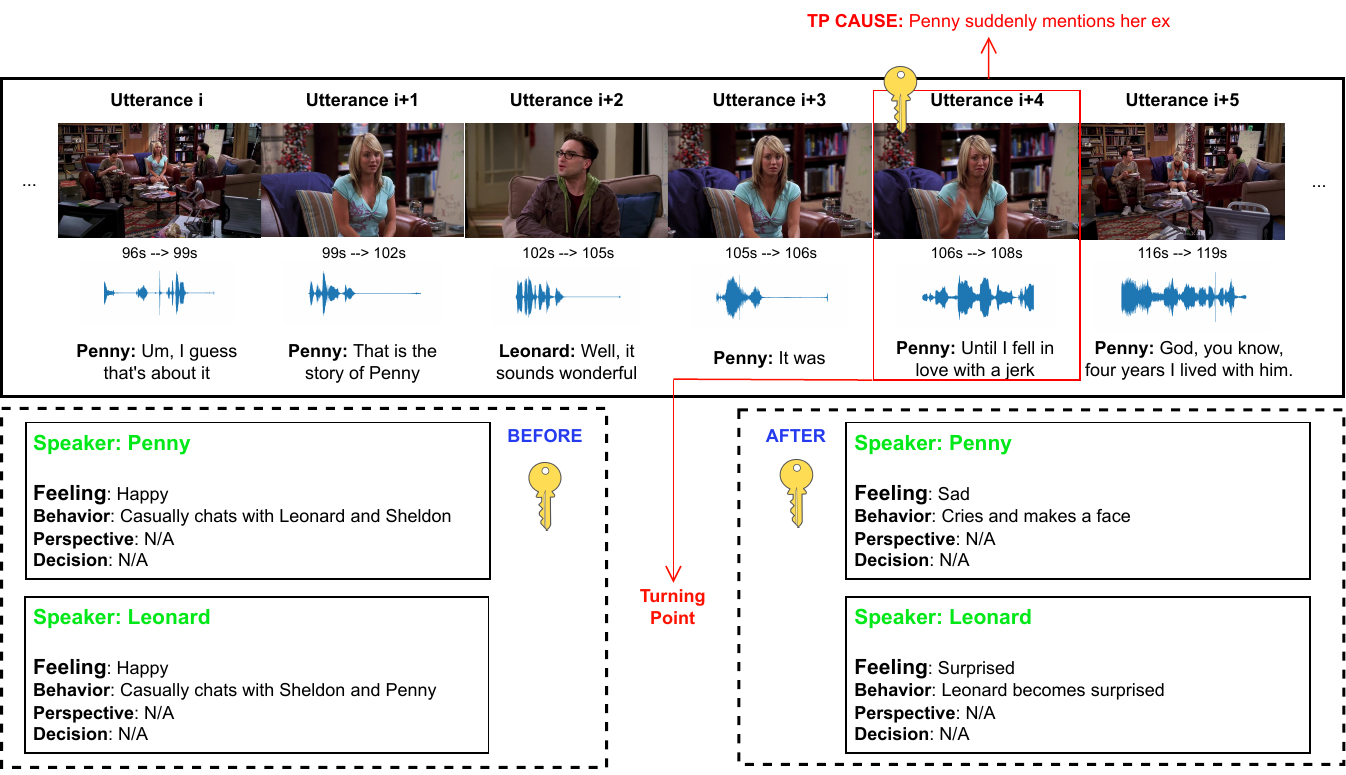} 
  \caption{Considering this example: Everyone is chatting casually. A turning point occurs when Penny (female character) starts crying, caused by her mentioning her ex while sharing her personal stories with Leonard and Sheldon (two male characters). According to human commonsense, this should be considered a significant change in the conversation because it catches the attention of the people watching, and the speakers involved (Leonard and Sheldon become confused).}
  \label{fig:MTPD_problem}
\end{figure*}

\section{Related work}
\label{sec: related work}
Multi-modal datasets have been developed for understanding human conversations
\cite{reece2023candor,meng2020openvidial,wang-etal-2023-vstar,firdaus-etal-2020-meisd,lei-etal-2018-tvqa,Li2023PTVDAL,shen2020memor}.
Each of them having limitations such as missing visual data, or providing just extracted features from it, missing context on shorter sequences, alignment issues and so forth.
To address these gaps, we developed a multi-modal conversational dataset from TV series episodes, featuring video content with timestamp annotations, aligned transcripts, and video frames, with annotations for turning points.

Turning points are a special case of change points \cite{aminikhanghahi2017survey} sometimes indicating a trend change direction or substantial change in intent for human data.
TPs in narrative analysis, as described by \cite{Keller:2020,papalampidi-etal-2019-movie,papalampidi2021movie}, denote critical moments that shape the plot and segment narratives into thematic units.
In psychology and social sciences, TPs are moments of significant change in individuals' perceptions, feelings, or life circumstances \cite{TurningPoints:2003,Wieslander2023TurningPA}. 
%
Our research adopts the TP definition from \cite{Keller:2020} and \cite{papalampidi-etal-2019-movie}, focusing on crucial moments within conversations that significantly impact discourse elements in human-simulated dialogues from a TV series. 
\citet{kumar2022discovering} introduces Emotion-Flip Reasoning (EFR), which is the task of identifying past utterances in a conversation that triggered a speaker's emotional state to change, aiming to explain emotional shifts during dialogue. For clarification regarding the differences, we not only provide information on emotional changes but also on the causes behind those changes. We specifically focus on significant emotional shifts. Moreover, we consider changes in decisions, perspectives, and behaviors as they are deemed significant. Additionally, we provide visual-textual evidence for these changes.


\section{Problem formulation}
\label{sec: problem}
The context of a casual conversation is denoted as \( C \), comprising \( m \) utterance-level videos \( U = \{u_1, \ldots, u_m\} \). Each utterance video \(u_i\) is associated with a corresponding text transcript and a speaker name \( \{t_i, s_i\} \).
We consider turning points within the conversation, in accordance with Definition \ref{def: turning point}. 

\begin{definition}
\label{def: turning point}
A \textbf{turning point} in this context is a moment that belongs to an utterance in a conversation, triggered by an identifiable event (that is called the turning point cause). This moment marks the beginning of unexpected or significant changes in the subjective personal states of at least one participant (such as decisions, behaviors, perspectives, and feelings) \footnote{We identified these states through a process of group discussions, video analysis, and literature review in Section \ref{sec: related work}, focusing on the most common variables in the post-analysis of casual conversations.}. We have annotated it with a timestamp and a textual explanation of its cause (Further elaboration on the definition is in appendix \ref{app:definition_elaboration}).
\end{definition}

Our proposed problem inputs consist of utterance-level videos with corresponding transcriptions, speaker names, and timestamps bound to the transcript. The problem can be divided into three tasks. The first task, referred to as MTPC (\textit{Multi-modal Turning Point Classification}), involves determining if a conversation includes any turning points (TP). The second task, MTPD (\textit{Multi-modal Turning Point Detection}), focuses on pinpointing the timestamps of these turning points in the conversations. A correct turning point is identified when the predicted timestamp falls within a time window threshold $\delta_t$ relative to the ground truth. The third task, MTPR (\textit{Multi-modal Turning Point Reasoning}), aims to discern the reasons behind each turning point, presented as a textual description. This task is crucial for formulating potential solutions to address negative turning points and gaining insights into cultural norms. Regarding evaluation, the model's timestamp predictions can be assessed qualitatively. However, we believe that the textual causes should be evaluated by human experts. Currently, we have not identified a qualitative method for evaluating textual causes, considering it as a potential avenue for future research.

\begin{table}[!ht]
    \small 
    \centering
    \begin{tabular}{|l|l|}
        \hline
        Total number of conversation videos & 340 \\
        Total duration (h) & 13.3 \\
        Total number of utterance-level videos & 12351 \\
        Total number of words in all transcripts & 81909 \\
        Average length of conversation transcripts & 241.5 \\
        Maximum length of conversation transcripts & 460 \\
        Average length of conversation videos (s) & 1.9 \\
        Maximum length of conversation videos (m) & 2.5 \\
        Total number of TPs videos & 214 \\
        \hline
    \end{tabular}
    \caption{Statistics of the MTP Dataset}
    \label{table:mtp-stats}
\end{table}

\begin{table*}[!ht]
\small
\centering
\begin{tabular}{|c|cccc|ccc|}
\hline
\textbf{Methods} & \multicolumn{4}{c|}{\textbf{Turning point classification}} & \multicolumn{3}{c|}{\textbf{Turning point detection}} \\ \cline{2-8} 
                                   & Precision     & Recall        & F1            & AUC  & Precision & Recall & F1   \\ \hline
\textbf{GPT-3.5}                   & 0.7           & 0.84          & 0.76          & 0.47 & 0.44      & 0.6    & 0.45 \\
\textbf{GPT-4}                     & \textbf{0.81} & \textbf{0.96} & \textbf{0.88} & 0.52 & 0.43      & 0.75   & 0.51 \\ \hline
\textbf{GPT-4 w/o tracking prompt} & 0.69          & 0.95          & 0.8           & 0.47 & 0.31      & 0.69   & 0.43 \\ \hline
\textbf{GPT-4 + few shot}         & 0.71       & 0.95       & 0.82       & \textbf{0.53}       & \textbf{0.52}    & \textbf{0.87}   & \textbf{0.61}   \\ \hline
\end{tabular}
\caption{Performance metrics for turning point classification and detection using different comparison methods}
\label{table:performance}
\end{table*}

\section{The MTP Dataset}
"The Big Bang Theory" \citep{bigbangtheory} provides a rich source of casual conversations, forming the foundation of our study. The eccentricities of its characters create a unique backdrop for sensitive moments crucial to our turning points analysis. Our three-stage process involves human annotators determining scene start and end times (Subsection~\ref{ssct:sbc}), extracting videos for conversations. The second phase (Subsection~\ref{ssct:culv}) annotates turning points based on guidelines explained in appendix~\ref{app:annotation}, while the third stage annotates relevant information, such as visual-textual evidence for observed changes.

\subsection{Scene boundary annotation}
\label{ssct:sbc}
Since an episode can contain multiple scenes, but our focus is solely on studying conversations within each scene, we conducted scene boundary annotation. In the first phase, we initiated scene boundary annotation by providing videos (crawled from the internet), scene's tags, and their initial sentences extracted from \citet{mitramir5_bigbangtranscript} to annotators. They were tasked with accurately identifying the start and end times of scenes by watching the videos and using the first sentences as cues as explained in annotation details in appendix~\ref{app:scene_boundary}. The statistics of the dataset can be found in Table \ref{table:mtp-stats}. 

\subsection{Creating utterance-level videos}
\label{ssct:culv}
WhisperX \citep{bain2023whisperx} was employed to segment conversation C into utterance-level videos ($U = \{u_1, \ldots, u_m\}$) with precise timestamps ($\delta T = \{\delta t_1, \ldots, \delta t_m\}$) and transcripts ($T = \{t_1, \ldots, t_m\}$). We found that the speaker identifier is crucial for human annotators to locate the turning points. To address this, we utilized an online dataset \citep{bain2023whisperx} containing speaker identifiers for Big Bang Theory episodes. Using GPT embedding search and the LLAMA model for prompting, we matched each utterance transcript $t_i$ to the corresponding speaker ID. Finally, human refinement was employed to ensure accurate alignment. This process resulted in triplets $\{t_i, \delta t_i, s_i\}$ for each utterance $u_i$ in conversation C, with $s_i$ representing the speaker for utterance $i$ (further details are provided in appendix~\ref{app:preprocessing}).

\subsection{Multi-modal Turning Point Annotation}
We assembled a team of three annotators, all of whom are proficient English-speaking students. Each conversation was then assigned to two annotators for annotation with clear guidelines (appendix~\ref{app:annotation}). The third annotator was designated as a judge responsible for reviewing the annotations and engaging in discussions with the first two annotators.

\subsection{Turning Point Evidence Annotation}
Once annotators identify turning points, they provide pre- and post-change details for a nuanced understanding. Clear explanations are required when annotators perceive no turning point, enhancing comprehension of situations considered unremarkable. Additionally, annotators timestamp moments of change in feelings, behaviors, decisions, and perspectives, substantiating observations with visual or verbal evidence.


\subsection{Feelings Annotation}
Annotators are asked to focus on emotions closely tied to turning points, ensuring clarity in decisions, behaviors, or perspectives before and after these moments. The incorporation of a feelings recognizer is motivated by recognizing emotions as vital markers in conversations. By highlighting feelings associated with turning points, annotators reveal emotional undercurrents shaping responses. We believe that proficient emotion recognition in the valence-arousal space aids in discerning significant changes in feelings, crucial for identifying turning points. However, due to resource constraints, we use common classes from the circumplex model of emotion \citep{russell1980circumplex} (see appendix \ref{app:feeling} for the model) instead of annotating valence and arousal for each emotion, enhancing precision and providing a structured framework for annotators to navigate human emotions systematically. An annotator selects frequent emotions from the circumplex model, defining a list including Positive (Happy, Excited, Calm, Relaxed, Alert), Negative (Anxious, Angry, Disgusted, Sad, Upset, Depressed, Frustrated, Confused), and Neutral/Transitional (Surprised, Neutral, Serious, Nervous) emotions.

\subsection{Annotation consensus}
After annotators completed their tasks, a group discussion session was organized to review and discuss conversation labels. The aim was to decide whether to keep, add, or delete turning points. This resulted in 340 conversations, with 214 having turning points and 126 without. Agreement was reached when annotators and the judge agreed on turning point labels, occurring in approximately 82\% of the dataset's turning point events. If all three annotators identify three distinct turning points (though this scenario didn't happen), the sample would be deleted due to the lack of unanimous agreement. Typically, we retain annotations receiving at least two out of three votes for a turning point. In our review session, when annotators identified the same turning points but provided different yet reasonable evidence, we merged their before and after evidence (including emotions and behaviors changes).

\section{TPMaven framework}
We present TPMaven, a language model prompting framework engineered to identify and ground turning points in casual conversational videos. The framework comprises two key components: 1) a scene describer that captures the visual information and articulates the essence of each utterance; and 2) a robust reasoner that interprets instructions, locating and elucidating turning points. For the first component, we prompt the LLAVA model \citep{liu2023llava} as our scene describer to get the relevant visual description of the scenes (frames) in the conversations. For the second, various ChatGPT models are prompted with a system prompt, including the definition of TP and three prompts for turning point identification: a describing instruction, the conversation $ C = \{<t_1, v_1, s_1>, \ldots,<t_m, v_m, s_m>\}$, with $v$ being the visual description, an optional tracking prompt to direct ChatGPT to track individual in the conversation, and a command prompt. Further details on the prompting templates for both components can be found in appendix~\ref{app:TPMAVEN}. 

\section{Experiments}
\label{sec: experiments}
We use LLAVA-7B \citep{touvron2023llama} to extract visual information in scene descriptions. GPT-3.5-1106 (a version of GPT-3.5 \citep{openai:chatgpt}) and GPT-4-1106 identify turning points, addressing context length issues. For assessing turning point localization, we focus on the positive set with 214 conversations. True positives are determined when predicted timestamps fall within $\delta_t$ = 20 seconds of ground-truth timestamps. During segmentation, we map GPT model outputs (utterance indices) back to timestamps for comparison (see more details in appendix~\ref{app:exp}). The performance metrics, including Precision, Recall, F1 and Area Under the Curve (AUC) are reported for each method in Table \ref{table:performance}. GPT-4, especially with few-shot learning, stands out as the most promising method for turning point classification, surpassing GPT-3.5 and GPT-4 without tracking prompts. We also found that the grounding output of GPT-4 is much concise in terms of tracking compared to other GPT models.

\section{Conclusion}
\label{sec: conclusion}
In conclusion, our research addresses the crucial task of recognizing pivotal moments in conversations, presenting a detailed taxonomy and a curated dataset called MTP. Our baseline framework, TPMaven, utilizes vision-language and GPT models for classification and detection, demonstrating its performance across various metrics. While TPMaven provides explainable predictions for sensitive moments, experimental results highlight the need to discern conversations with and without turning points. Future directions are in appendix~\ref{app:future}.

\section*{Limitations}
The dataset is designed for post-analysis to understand what captures the attention of viewers in videos and speakers during conversations. Due to resource limitations, we could only curate a single-lingual dataset focused on critical moments in English culture. Unfortunately, we had to opt for simple emotion annotation instead of the more informative valence-arousal space annotation, which would provide intensity and direction of emotions.

Furthermore, we faced challenges in evaluating the Multi-modal turning point reasoning task. While attempting to utilize another GPT-4 as an evaluator for explanations on some samples, followed by human verification, we encountered inconsistent results. Despite our belief that human evaluation is optimal, resource constraints prevented us from pursuing this approach. Emotion reasoning was excluded for the same reason.

Regarding scene-describing methods, we have employed LLAVA due to its cost-effectiveness. Although a faster version of GPT-4 was available \citep{gpt4vision} during the submission of this work, which could potentially improve scene descriptions, budget limitations hindered us from exploring its use.

In this problem, the input should simply be a video, and the output should consist of the turning points. However, at the time of conducting this research, we have not identified any reliable speaker identification method; therefore, this aspect may be addressed in our future research. As speaker IDs are crucial for tracking the states of each individual in the conversation, and it is reasonable to assume that speakers are known through the normal mental human annotation process, we believe it is justifiable to human-annotate that information instead of relying on an inaccurate speaker ID. The latter could lead to expected underperformance. It is important to note that turning points should also encompass non-verbal cues. Currently, we only consider verbal turning points that occur within an utterance. The case of online turning point detection, where turning points are identified in real-time, has not been explored in our research at this time. Additionally, we believe that the definition of a turning point can be broadened to encompass specific conversational contexts beyond casual discourse, such as political discussions. In these situations, even slight changes in subjective states can lead to significant norm violations. Conversely, in our scenario of casual conversations among friends, a much higher threshold should be considered to distinguish between meaningful event changes and insignificant ones.

\section*{Ethics consideration}
\textbf{Data life-cycle and access}: Our dataset has been scrutinized and approved by the relevant institutional committees. All annotators have agreed to relevant terms and participated in training sessions. They were compensated at a rate significantly higher than the local minimum wage. The resources presented in this work are utilized for research purposes only. We have obtained all data copyrights pertinent to this paper. To ensure proper citation and prevent malicious application, we have prepared detailed instructions, licenses, and a data usage agreement document that we link in our project repository. Additionally, we intend to make our software available as open source for public auditing.

\textbf{Copyrights}
Our dataset incorporates videos from 'The Big Bang Theory' television series for training AI models in natural language understanding tasks. The inclusion of copyrighted material raises important considerations regarding fair use and transformative use under copyright law. We assert that our use of these videos qualifies as fair use, as it is conducted for transformative purposes aimed at advancing scientific understanding and innovation. Specifically, our research involves the transformation of the original videos through linguistic analysis and modeling, contributing novel insights into conversational comprehension. Furthermore, our use of the videos is limited in scope and does not detract from the commercial market for the series. We provide appropriate attribution to the copyright owner of the show and take measures to ensure that the dataset is used responsibly and ethically within the research community.

\textbf{Data bias}: When pinpointing a crucial turning point, the evidence reflecting subjective personal states (feelings, behaviors, perspectives, decisions) may exhibit variations. Annotators, expressing diverse viewpoints on the same event in human language, can contribute to this divergence. Consequently, the explanations and evidence surrounding the turning point may incorporate personal bias in articulating the matter. We advise future users of the dataset to be mindful of this potential bias.

\section*{Acknowledgements}
This research is based upon work supported by U.S. DARPA under agreement No.~HR001122C0029. The opinions, views, and conclusions contained herein are those of the authors and should not be interpreted as necessarily representing the official policies, either expressed or implied, of DARPA or the U.S. Government. The U.S. Government is authorized to reproduce and distribute reprints for governmental purposes, notwithstanding any copyright annotation therein. We appreciate all annotators for their contributions to this work.
We would also like to thank Prof.\ Heng Ji for her valuable feedback.


\bibliographystyle{acl_natbib}

\appendix
\section{MTP Dataset creation details}
\subsection{Preprocessing}
\label{app:preprocessing}

In analyzing conversation C, we utilized WhisperX \citep{bain2023whisperx} to segment each video into $m$ utterance-level videos ($U = \{u_1, \ldots, u_m\}$) with precise start and end timestamps ($\delta T = {\delta t_1, \ldots, \delta t_m}$) for each transcript ($T = \{t_1, \ldots, t_m\}$). 

Speaker IDs for each utterance were annotated by a process of matching with the transcripts and speaker labels from the scenes in \citet{mitramir5_bigbangtranscript}. For each utterance extracted by WhisperX, we need to find the row in \citet{mitramir5_bigbangtranscript} to extract the speaker name. This can be done by matching the corresponding transcript from WhisperX and the row from \citet{mitramir5_bigbangtranscript}. Using GPT-3.5, we created an embedding file for each scene extracted from \citet{mitramir5_bigbangtranscript}, where each line represents a text pair of utterance and corresponding speaker ($u', s$). Through an embedding search for each WhisperX-extracted utterance $u_i$, we retrieved the most similar sentence $u'_i$ from the pre-processed \citet{mitramir5_bigbangtranscript} with its corresponding speaker $s_i$. We prompted LLAMA-7b with transcript $t_i$ and the candidate sentence, including speaker names from the search model, to assign the speaker for each utterance. Recognizing potential unintended outputs from LLMs, human annotators meticulously verified speaker identification, ensuring accurate alignment with respective names in the transcripts.

\subsection{Annotation}
\subsubsection{Scene Boundary}
\label{app:scene_boundary}
It is crucial to emphasize that our episodes consist of various scenes and transitions, requiring the annotation of scene boundaries. To streamline this task, we enlisted a team of students to view the videos. They were tasked with assigning scene tags and providing the initial sentence for each scene, serving as a prompt to expedite the process. This meticulous process resulted in the identification of 340 conversations, comprising a comprehensive 13.3 hours of video content for our study.

\subsubsection{Turning Points}
An example of our turning point annotation can be found in Table \ref{sample}.

\begin{table}[!ht]
    \small 
    \centering
    \begin{tabular}{|p{0.3\linewidth}|p{0.65\linewidth}|}
        \hline
        \textbf{scene} & A corridor at a sperm bank. \\
        \hline
        \textbf{duration} & 150 \\
        \hline
        \textbf{conversation} & 1 \\
        \hline
        \textbf{TP\_location} & 01:25 \\
        \hline
        \textbf{TP\_cause} & Sheldon shows his concerns about donating sperm \\
        \hline
        \textbf{pre\_point\_feeling} & neutral (1:24) \\
        \hline
        \textbf{post\_point\_feeling} & nervous (1:38) \\
        \hline
        \textbf{pre\_point\_dbp} & Leonard and Sheldon plan to donate sperms so that they can have extra money (1:45) \\
        \hline
        \textbf{post\_point\_dbp} & Leonard and Sheldon leave the room (2:29) \\
        \hline
        \textbf{explanation} & According to commonsense, there is a clear change in their decisions. \\
        \hline
    \end{tabular}
    \caption{A sample turning point annotation for conversation 1 in our dataset. \textbf{pre\_point\_dbp} and \textbf{post\_point\_dbp} stands for pre-point and post-point decisions, behaviors, perspectives respectively.}
    \label{sample}
\end{table}

\subsubsection{Feelings}
\label{app:feeling}
\begin{figure}
  \centering
  \includegraphics[width=\columnwidth]{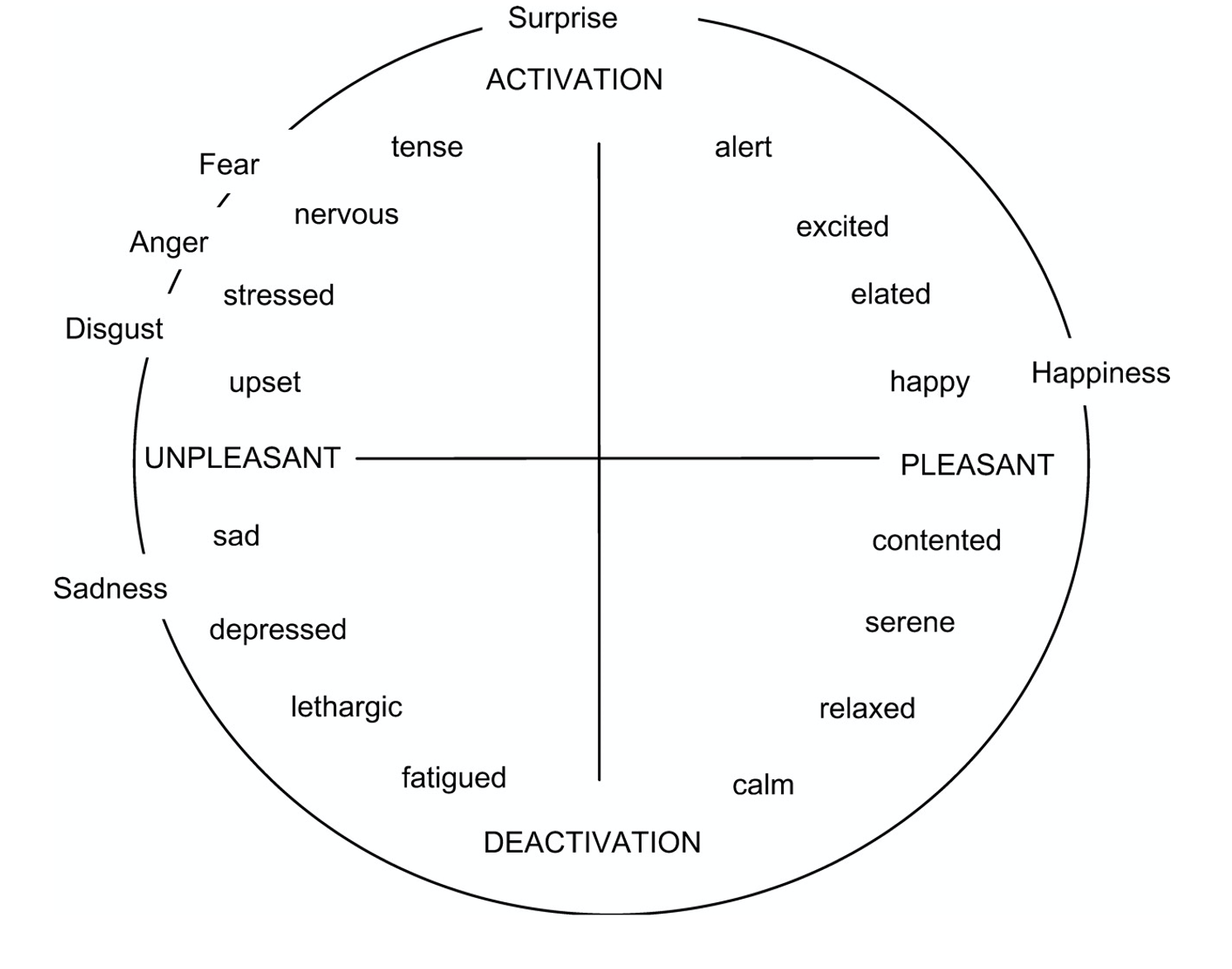} 
  \caption{The circumplex model of emotions in \citep{russell1980circumplex}}
  \label{fig:circumplex_model}
\end{figure}

Annotators are asked to focus on emotions closely tied to the turning points, ensuring clarity in decisions, behaviors, or perspectives before and after these turning points. The intuition behind incorporating a feelings recognizer lies in the recognition that emotions serve as vital markers of key moments in a conversation. By focusing on feelings closely associated with turning points, annotators can illuminate the emotional undercurrents that shape individuals' responses and reactions. For instance, someone may say something offensive, but whether it forms a turning point depends on the other person's reactions. We also believe that a proficient emotion recognizer within the valence-arousal space proves valuable in discerning significant changes in feelings. Without knowing the intensity and direction of these changes, identifying turning points becomes challenging. To avoid overcomplicating the annotation process due to resource constraints, we opt for common classes in the circumplex model of emotion depicted in Figure \ref{fig:circumplex_model} instead of annotating valence and arousal for each emotion. The circumplex model of emotion enhances this process by providing a structured dimension. This model maps emotions based on underlying dimensions such as valence and arousal, ensuring systematic classification. It not only enhances labeling precision but also offers annotators a practical framework to navigate the intricate landscape of human emotions.

\subsection{Statistics}
\subsubsection{Different types of turning points}
After annotating the data, we provide ChatGPT with all the causes of turning points and categorize the types in Table \ref{tab:tp_types}.
\begin{table}[!ht]
\centering
\begin{tabular}{|p{0.25\columnwidth}|p{0.65\columnwidth}|}
\hline
\textbf{Types} & \textbf{Explanation} \\
\hline
Emotional Outbursts & Sometimes, when someone gets really, really mad and can't control it, it can lead to a big, angry fight. \\
Changes in Decisions & Sometimes, the group has a plan, but suddenly they decide to do something different. \\
External Influences & Imagine someone new joins the conversation, and it completely changes how everyone feels or what they think. \\
Shifts in Perspective & Sometimes, everyone starts thinking one way, but later on, they change their minds and think differently. \\
Uncomfort-able Situations & Imagine someone violating social norms, and it makes everyone feel uncomfortable or upset. \\
\hline
No Turning Points &
\begin{tabular}[t]{@{}p{0.65\columnwidth}@{}}
- Even when someone says something mean, everyone reacts like they normally would, without any big changes. \\
- Sometimes, during the conversation, nobody's subjective personal states change much; things stay pretty much the same.
\end{tabular} \\
\hline
\end{tabular}
\caption{Different categories of turning points (TP) types were identified by prompting and providing ChatGPT with a list of TP causes from our dataset.}
\label{tab:tp_types}
\end{table}

\subsubsection{Emotional shifts}
\begin{figure*}[!ht]
  \centering
  \includegraphics[width=\textwidth]{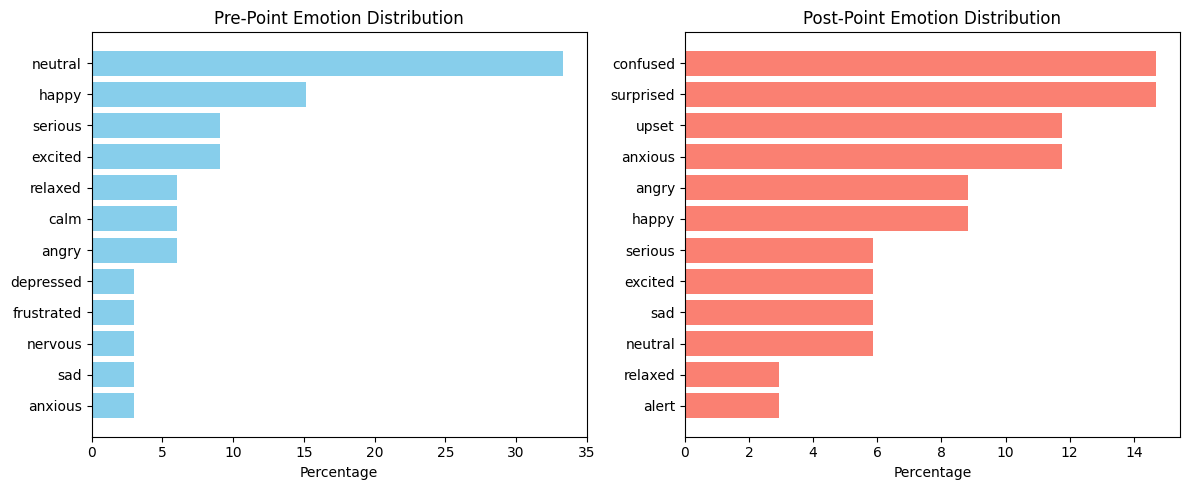} 
  \caption{Emotional distribution of the top 20 most occurrences before and after the turning point in our dataset. This caption summarizes the analysis of emotions in relation to the most frequent occurrences, highlighting changes around the identified turning point in the dataset.}
  \label{fig:feelings_stat}
\end{figure*}

We also provide the analysis of the most common types of emotional changes before and after turning points in Figure \ref{fig:feelings_stat}.

\section{Turning points annotation guidelines}
\label{app:annotation}
\subsection{Further elaboration on the definition}
\label{app:definition_elaboration}
Considering definition \ref{def: turning point}, we want to elaborate some important terms.

\subsubsection{The term ``\textit{identifiable}''}
This means the \textbf{event} can be recognized based on clear evidence.

\begin{table}[!ht]
    \small 
    \centering
    \begin{tabular}{|l|}
        \hline
        Leonard: Penny’s up.                           \\
        Penny: You sick, geeky bastards!               \\
        Leonard: How did she know it was us?           \\
        Sheldon: I may have left a suggested organizational \\ 
        schematic for her bedroom closet. \\
        Penny: Leonard!                                \\
        Leonard: God, this is going to be bad.         \\
        Sheldon: Goodbye, Honey Puffs, hello Big Bran. \\
        Penny: You came into my apartment \\ 
        last night when I was sleeping?          \\
        Leonard: Yes, but, only to clean.              \\ \hline
    \end{tabular}
    \caption{A sample transcript of a conversation in our dataset}
    \label{table:conversation-sample}
\end{table}

Considering a conversation from Table \ref{table:conversation-sample}, the identifiable events are:
\begin{enumerate}
  \item Penny discovers Leonard and Sheldon entering Penny's apartment and confronts them about it.
  \item Leonard and Sheldon try to explain their actions and justify themselves.
\end{enumerate}

\subsubsection{The term ``\textit{subjective personal states}''}
These encompass changes in a speaker's:

\begin{itemize}
  \item \textbf{Decisions:} Choices made during the conversation.
  \item \textbf{Behaviors:} Actions taken during the conversation.
  \item \textbf{Perspectives:} Shifts in the way a speaker sees or understands a topic.
  \item \textbf{Feelings:} Emotional states.
\end{itemize}

\subsubsection{The term ``\textit{Unexpected}''}
The event should be surprising and deviate from the usual flow or expectations of the conversation.

\subsubsection{The term ``\textit{Significant}''}
The change should be of significance, impacting not only the individual but also affecting the dynamics of the conversation.

\begin{itemize}
  \item It affects not only one person but also those around them.
    \begin{itemize}
      \item Example: When Person A cries, it makes Person B cry too.
    \end{itemize}
  \item The impact on the subjective personal states can differ, but it should make common sense.
    \begin{itemize}
      \item Example: Changing your mind from staying in to going out is considered significant.
      \item Example: Changes in how you act, like going from being neutral to getting into a debate or becoming more engaged, are considered significant.
      \item Example: Going from feeling normal to feeling heartbroken is considered significant.
    \end{itemize}
\end{itemize}

\subsubsection{The term ``\textit{During}''}
The annotators are asked to consider the evidence before and after that point in the current conversation only, not the potential consequences.

\subsubsection{The goal of detecting TPs}
In healthcare monitoring, we have two scenarios. For critical patients, we use a low sensitivity threshold to detect even subtle changes due to their sensitivity. For general patients, we employ a high sensitivity threshold to identify only the most significant changes, avoiding unnecessary alerts.

Similar to general patient monitoring, our research objective is to identify important moments in casual conversations. We focus on recognizing changes that match our definition of significance while ignoring minor ones. This knowledge base serves as a valuable resource for developing applications, encompassing conversation analysis to mitigate miscommunication, study decision-making, and behaviors, and highlight key aspects of conversations.

\subsection{Annotation Flows}
The annotators are given a video of a conversation and asked to follow three phases of annotation.
\subsubsection{First phase}

In this initial phase, understand the content and flow of the conversation. Identify the topics, speakers, and main events without focusing on turning points.

\subsubsection{Second phase}

The annotators are asked to find an event in the conversation that causes a turning point, and then label the timestamps where the change occurs. There can be multiple turning points.

\paragraph*{Recommended Steps:}

\begin{enumerate}
    \item Evaluate each speaker separately.
    \item Analyze changes in decisions, behaviors, perspectives, and feelings independently.
    \item If a change meets the criteria of being \textbf{significant} and \textbf{unexpected}, mark the timestamp when the change starts. Also, write down a short summary of the event that started the change (the cause of the turning point).
    
    The change in the subjective personal states of a person can be caused by that person or another person, you should write down the event that caused the turning point (\textbf{who does what}). If it is caused by a person himself (by re-thinking, etc.), you should write down something like "Penny realizes that ..." or "Sheldon decides to ..."
    
    \item Please note the changes both before and after the turning point. While changes in decisions, behaviors, and perspectives are typically evident, when it comes to feelings, concentrate only on those that are closely linked to the turning point. The person whose subjective personal states change will have a clear pre-point and post-point decision or behavior or perspective. You should write who does what too. Additionally, if there is a change in feelings but no corresponding change in decisions, behaviors, or perspectives, please provide a clear explanation of why that change is significant. Since human emotions can change frequently, our focus should be on reasonably significant emotional changes within that context. 
    
    \item Mark the timestamp for the evidence associated with those changes in parentheses. The evidence can consist of verbal or non-verbal cues. For example, 'sad (1:05)' indicates that the evidence is located at 1 minute and 5 seconds into the video. At 1:05, a person might say something like, “I broke up with my girlfriend,” which provides strong evidence of the feeling of sadness. Alternatively, at 1:05, there is a frame capturing his sadness expressed through his facial expressions.

\end{enumerate}

\textbf{Key Guidelines}
\begin{itemize}
    \item Decisions, behaviors, and perspectives are more likely to trigger a turning point, as it is defined to capture decisive moments in a conversation.
    \item When it comes to feelings, it's important to consider the context of why and how they change. This helps us conclude whether there's a significant shift influencing the emotional dynamics of the conversation.
    \item Ensure turning points are clear and memorable, leaving a lasting impression.
    \item If no significant moment is found in the first two phases, move on to the next conversation.
    \item Envision yourself as an impartial observer to identify surprising or attention-grabbing moments.
    \item Focus on sudden reactions indicating a noteworthy change in the casual conversation dynamics.
    \item Approach each video with fresh eyes, treating characters as unfamiliar individuals.
\end{itemize}

\subsubsection{Third phase}
If a point is labeled as a turning point and you believe it is not adequately represented by the pre-point, post-point, and TP\_cause columns, please comment on the additional evidence you think is necessary for a conclusive determination.

If you are uncertain whether it qualifies as a turning point, provide a clear explanation, and express any concerns you may have.



\section{TPMaven framework}
\label{app:TPMAVEN}
We present TPMaven, a language model prompting framework engineered to identify and ground turning points in casual conversational videos. The framework comprises two key components: 1) a scene describer that captures and articulates the essence of each utterance, providing a comprehensive understanding of the visual information; and 2) a robust reasoner that interprets instructions, skillfully locating and elucidating turning points, offering insightful explanations for shifts in the conversation. 

\subsection{Scene describer}
Originally, our intention was to utilize the video-language understanding model Video-LLAMA. However, due to prolonged processing times, we opted for an expedited alternative, extracting a list of frames denoted as $F = \{f_1, \ldots, f_m\}$, wherein each frame corresponds to an individual utterance.

To expedite the process, we opted for LLAVA, a vision-language model that demonstrated satisfactory results in human evaluations and improved processing efficiency compared to Video-LLAMA. While GPT-4 integrated with images was considered, it was dismissed due to cost constraints. Subsequently, each utterance in the video is now denoted by a paired set $\{t, f\}$, where $t$ signifies the transcript, and $f$ represents a randomly selected frame during that utterance. Given that TV series consistently feature the speaker's face in every utterance, selecting a random frame serves as a sufficient baseline for capturing visual information. This approach is also computationally efficient.

The examination of visual stimuli within conversations yields rich evidentiary material, encompassing facial expressions and behavioral cues. These visual indicators are instrumental in constructing a comprehensive narrative of the discourse. Hence, we use this prompt: \textit{``Give me the short descriptions of the actions, facial expressions, postures, gestures, potential emotions (with valence and arousal)''} to retrieve the relevant information (including actions and affective factors) that can help us to detect the turning points. 

Given the verbosity of LLAVA's outputs and its potential impact on the context length of the GPT model, we employ a GPT-3.5 model for summarization. Eventually, we get a set of visual description for each utterance in the conversational 



\subsection{Reasoner}
Pretrained language models (PLMs) store implicit knowledge about the world learnt from large-scale text collected around the internet \citep{petroni-etal-2019-language}. There has also been previous attempts to use LLMs as a reasoner for a variety of tasks \citep{kojima2022large}. Our hypothesis is that if we are efficient at telling the story of the conversation to the LLMs and inspired from the CoT methods, if we can prompt a series of relevant prompt that can lead and guide the LLMs towards answering basic questions that it is trained on and is having in its internal knowledge, it can produce desireable results. Thus, we strive to break our tasks down. 

From the above steps, each conversation $C$ consists of $m$ utterances can now be represented as $ C = \{<t_1, v_1, s_1>, \ldots\,<t_m, v_m, s_m> \}$ with $t_i$, $v_i$ and $s_i$ being the transcript, visual description and speaker for an utterance $i$ respectively. 
Our prompting template concatenates multiple sub components prompts, each with its own functionality in guiding the LLM:

\begin{itemize}

\item \textbf{describing\_instruction} - \textit{``Read this conversation. Each utterance includes the transcripts and visual descriptions.''} - This is followed by filling the conversation in the form of a set of utterances U. 

\item \textbf{tracking\_instruction} - \textit{``Utilize a tracker for each person in the conversation. For each speaker, provide a concise list of their feelings, behaviors (based on the context and actions), decisions, and any perspective changes (include those with clear evidence from the conversation). Limit the list to a maximum of 256 words.''}

\item \textbf{commanding\_instruction} - \textit{``Identify the turning point events based on the initial conversation and track results if there are any. Begin by finding the turning point for each person.''}

\end{itemize}

We also leverage the system role in the ChatGPT Completion API, which is the role that helps provide fixed high-level instructions to the whole system, by filling in the \textbf{system\_content} field with this description: \textit{``You are a trained chatbot that can find turning points in conversations. A turning point in a conversation is an identifiable event that leads to an unexpected and significant transformation in the subjective personal states (including decisions, behaviors, perspectives, and feelings) of at least one speaker during the given conversation.''} - This prompt is used to fill in the \textbf{system\_content} of the ChatGPT completion API.

\subsection{Conclusion module}
We provide GPT-4 with this prompt: ``\textit{For each found turning point in the prediction, find the starting utterance index only. Return a list of n utterance start indices corresponding to a turning point in the prediction. Follow strictly this format in your response: e.g. utterances = [utterance\_5, utterance\_25]. Return None if there is no turning point found. Limit the response to 50 words.}'' and the conversation with utterance indices to retrieve the utterance indices that has turning points. Subsequently, we match these indices back to timestamps extracted in the pre-processing stage to compare with the timestamps' label.

\section{Experimental settings}
\label{app:exp}

\subsection{Implementation details}
For the scene describer, we utilize LLAVA-7B to extract visual information from an image. In the reasoning process, we leverage GPT-3.5-1106 and GPT-4-1106 versions to identify turning points. This choice is motivated by the large input size, mitigating potential context length issues encountered in conventional GPT turbo models from OpenAI. For the classification task, our primary evaluation metrics include Precision, Recall, and F1. Given the dataset's imbalance, we also incorporate the use of AUC. In the detection task, we focus on metrics such as P, R, and F1. To assess the performance of localizing turning points, we exclusively consider the positive set, comprising 214 conversations for evaluation. For each conversation, $k$ turning points are detected by TPMaven.  A true positive is determined if, for each ground-truth in the conversation, there exists a predicted timestamp falling within $\delta_t$ = 20 seconds. This is done as the turning point event found by ChatGPT can belong to several consecutive sentences. Since the GPT model's output from the conclusion module consists of a list of utterance indices, we map it back to the timestamp from the utterance-level segmentation phase for comparison.

\subsection{Discussion of the tracking prompts}
\begin{figure}[!ht]
  \centering
  \includegraphics[width=0.9\linewidth]{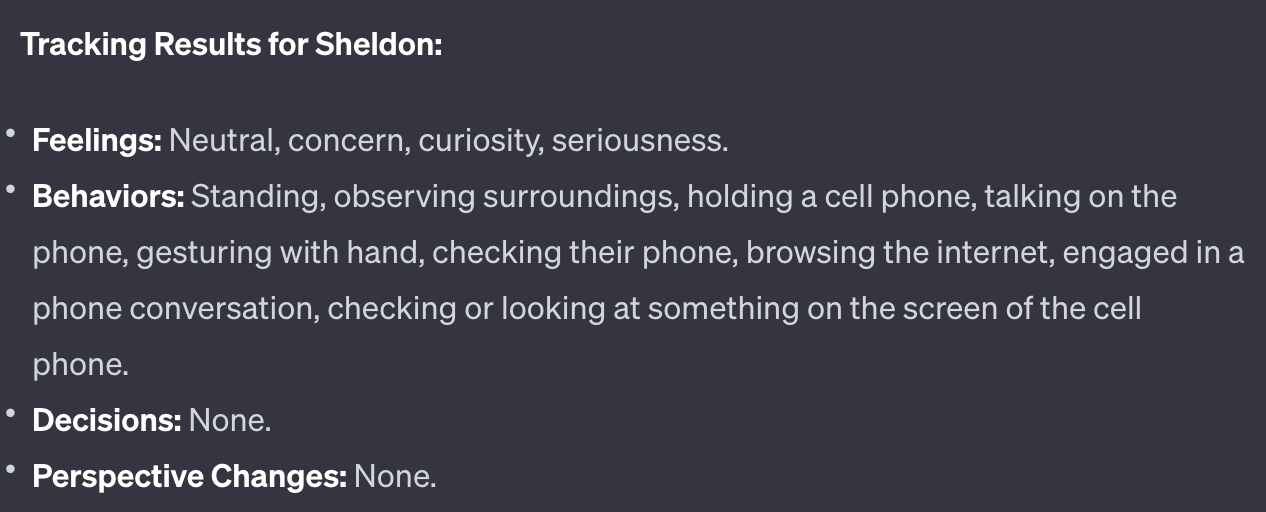} 
  \caption{Tracking results using GPT-3.5}
  \label{fig:gpt35}
\end{figure}

\begin{figure}[!ht]
  \centering
  \includegraphics[width=0.9\linewidth]{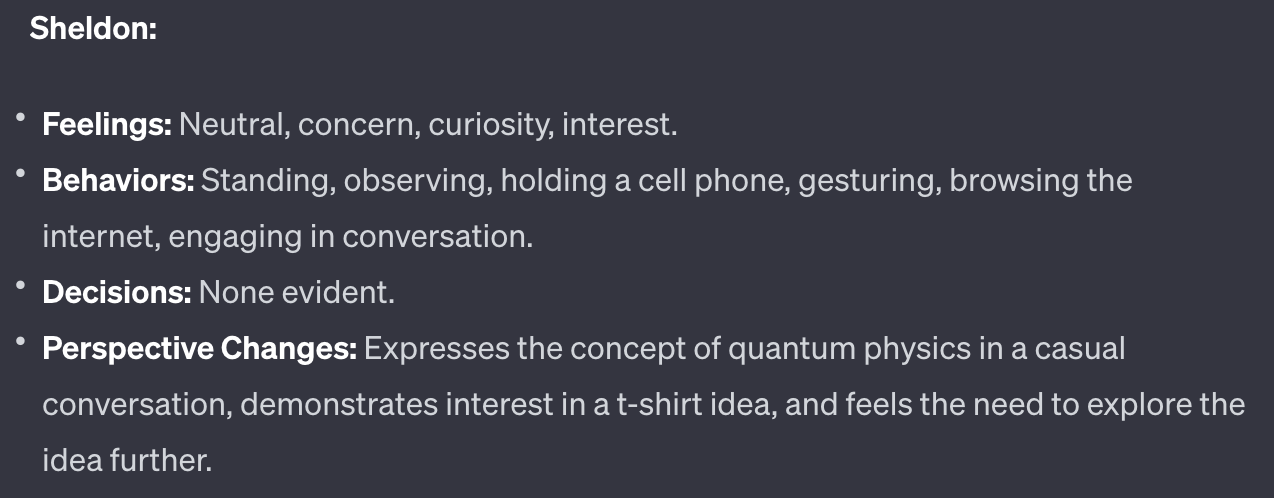} 
  \caption{Tracking results using GPT-3.5-turbo}
  \label{fig:gpt35turbo}
\end{figure}

\begin{figure}[!ht]
  \centering
  \includegraphics[width=0.9\linewidth]{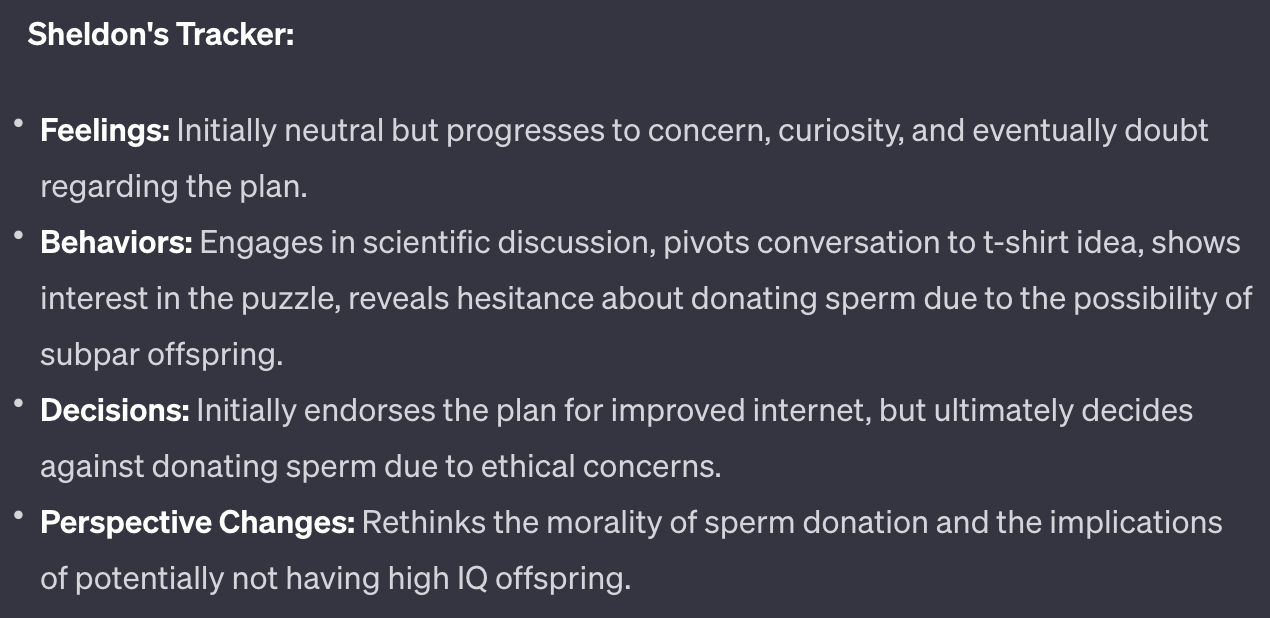} 
  \caption{Tracking results using GPT-4}
  \label{fig:gpt4}
\end{figure}

Given the conversation video between Sheldon and Leonard in the first scene of the series (Season 01, Episode 01) \citep{bigbangtheory} (Please refer to our project website to watch the video\footnote{\url{https://giaabaoo.github.io/TPD_website/}}), different GPTs are utilized with the tracking prompt. The results are depicted in Figure \ref{fig:gpt35}, \ref{fig:gpt35turbo} and \ref{fig:gpt4}. 

\section{Discussing future works}
\label{app:future}

In the course of conducting this research, we have identified several critical challenges that we believe are essential to address in future research on Multi-modal turning point detection. The following areas present promising avenues for further exploration:

\subsection*{Multi-lingual Multi-cultural Dataset}
Addressing the nuances in conversations across different languages and cultures, where norms vary, requires the development of a comprehensive multi-lingual, multi-cultural dataset. Such a dataset would capture the intricacies inherent in linguistic and cultural differences.

\subsection*{Emotion Recognition in Valence-Arousal Space}
The development of an effective emotion recognizer in the valence-arousal space holds the potential to enhance traditional time-series change point detection methods. Accurately identifying emotional shifts can contribute to the identification of candidate turning points.

\subsection*{Multi-modal Emotion Reasoning}
Our dataset not only captures turning points but also annotates changes in emotions related to these points. Therefore, there is an opportunity to develop methods in emotion reasoning using this dataset.

\subsection*{Multi-modal Turning Point Reasoning}
Providing the cause of the turning point and a causal chain of events related to feelings, behaviors, decisions, perspectives, etc., enables the development of a method or benchmark for turning point reasoning. However, a significant challenge lies in constructing a reliable evaluator to compare textual predictions from a model with the ground-truth explanations of turning points.

\end{document}